\documentclass[11pt]{article}

\usepackage[preprint]{WenyanGPT}

\usepackage{times}
\usepackage{latexsym}

\usepackage[T1]{fontenc}
\usepackage[utf8]{inputenc}

\usepackage{microtype}

\usepackage{inconsolata}
\usepackage{natbib}
\usepackage{graphicx}

\usepackage{CJKutf8}  % 传统方式支持中文
\usepackage{makecell}  % 表格中支持换行、合并单元格等操作

\usepackage[utf8]{inputenc}  % 支持 UTF-8 编码
\usepackage[T1]{fontenc}      % 使用 T1 编码，避免中文引号等问题
\nolinenumbers
\title{WenyanGPT: A Large Language Model for Classical Chinese Tasks}

\author{
  \textbf{Xinyu Yao\textsuperscript{1}},
  \textbf{Mengdi Wang\textsuperscript{1}},
  \textbf{Bo Chen\textsuperscript{1,2}},
  \textbf{Xiaobing Zhao\textsuperscript{1,2}}
\\
\\
 \textsuperscript{1}Minzu University of China,
 \\
\\
 \textsuperscript{2}National Language Resources Monitoring \& Research Center of Languages
\\
\\
\small{
    \href{mailto:email@domain}{chenbomuc@muc.edu.cn}
  }
}
\renewcommand{\abstractname}{Abstract}

\begin{document}
\maketitle

\begin{abstract}    
	Classical Chinese, as the core carrier of Chinese culture, plays a crucial role in the inheritance and study of ancient literature. However, existing natural language processing models primarily optimize for Modern Chinese, resulting in inadequate performance on Classical Chinese. This paper presents a comprehensive solution for Classical Chinese language processing. By continuing pre-training and instruction fine-tuning on the LLaMA3-8B-Chinese model, we construct a large language model, WenyanGPT\footnote{\url{https://huggingface.co/Wenyanmuc/WenyanGPT}}, which is specifically designed for Classical Chinese tasks. Additionally, we develop an evaluation benchmark dataset, WenyanBENCH\footnote{\url{ https://github.com/Wenyanmuc/WenyanBENCH}}. Experimental results on WenyanBENCH demonstrate that WenyanGPT significantly outperforms current advanced LLMs in various Classical Chinese tasks. We make the model's training data, instruction fine-tuning data\footnote{\url{https://github.com/Wenyanmuc/WenyanGPT}}, and evaluation benchmark dataset publicly available to promote further research and development in the field of Classical Chinese processing.
\end{abstract}

\section{Introduction}

Classical Chinese is an important component of Chinese culture, with a long history and profound cultural heritage. It is essential for understanding traditional Chinese culture. With the rapid development of artificial intelligence technology, intelligent processing of ancient Chinese texts offers a new solution for the preservation and inheritance of Classical Chinese. Modern technologies, such as digitalization and natural language processing, can efficiently preserve and spread traditional culture while fostering deeper and more innovative academic research. Maximizing the potential of AI in processing Classical Chinese texts has become an urgent need for cultural inheritance and academic development.

\begin{figure}
	\centering
	\includegraphics[width=0.5\textwidth]{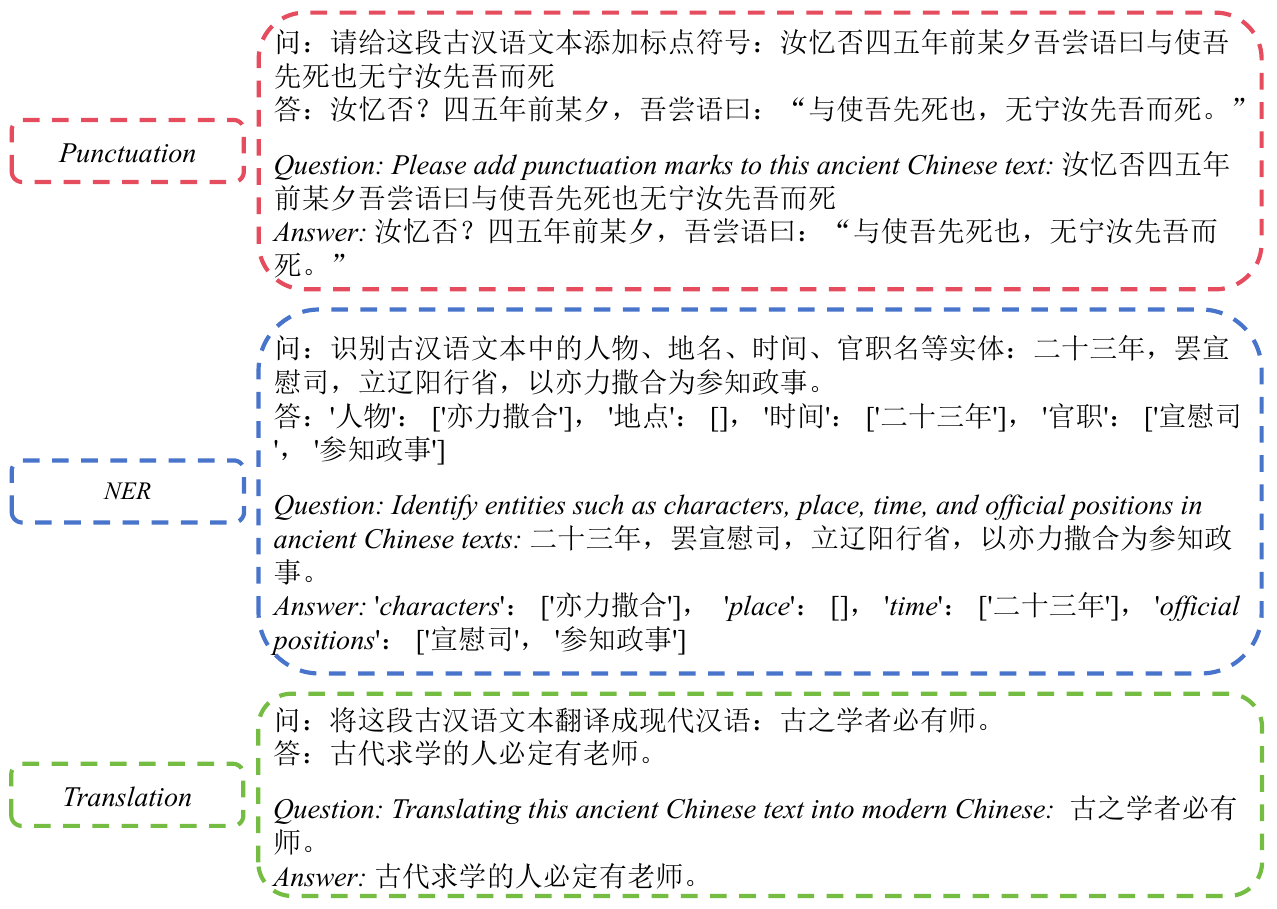}
	\caption{Examples of tasks from WenyanGPT. The model demonstrates advanced knowledge in Classical Chinese and shows strong performance in both Classical Chinese understanding and generation tasks.}
	\label{fig:1}
\end{figure}

Early research in Classical Chinese language processing focused on tasks such as punctuation, word segmentation, part-of-speech tagging, named entity recognition and translation. These tasks initially relied on traditional machine learning methods, such as Hidden Markov Models (HMMs) for part-of-speech tagging \cite{10.5555/647240.718633}, Conditional Random Fields (CRFs) for punctuation \cite{DBLP:conf/acl-sighan/HuangSC10} and named entity recognition \cite{XDTQ201903006,li2018automatic}. In deep learning, RNNs, LSTMs, GRUs, and Attention mechanisms have been applied to various tasks, including couplet generation and classical poetry generation \cite{yan2016chinese,yi2017generating}, punctuation and part-of-speech tagging using BiLSTM-CRF models \cite{wang2019ancient,cheng2020integration,zhang2023comparative,XDTQ202411008}. As Transformer architectures emerged \cite{Vaswani2017AttentionIA}, studies began using large-scale parallel corpora to train models for translating Classical Chinese into Modern Chinese \cite{Liu2018AncientModernCT} and for generating classical poetry \cite{huang2020generating}. The introduction of pre-trained models including BERT \cite{kenton2019bert} and GPT \cite{Radford2018ImprovingLU} provided new opportunities for intelligent Classical Chinese processing. Some research integrated ancient Chinese texts into the training data of general pre-trained models, improving the processing performance of Classical Chinese compared to typical pre-trained models \cite{Tian2020AnchiBERTAP,TSGL202206005,wang2023gujibert,Liu2023SikuGPTAG}. Other studies used Classical Chinese corpus to continue pre-training and fine-tuning large language models, aiming to build conversational models for Classical Chinese \cite{TSGL202410011,yang2024zhongjing,Cao2023TranslatingAC,Cao2024TongGuMC}.

However, challenges remain in Classical Chinese processing. Different tasks require training specialized models, and no effective universal model exists. Additionally, there is a lack of standardized evaluation benchmarks in this field, with existing evaluation tasks, datasets, and metrics being inconsistent, making it difficult to perform cross-task comparisons and systematic assessments of model performance.

To address these issues, we propose WenyanGPT, a large language model for Classical Chinese. Some examples of WenyanGPT are shown in Figure \ref{fig:1}. We also construct the largest available pre-training corpus for continued pre-training, enhancing the model's domain adaptability. Additionally, we propose a framework for generating domain-specific instruction data for supervised fine-tuning in the development of WenyanGPT. To promote research in the intelligent processing of Classical Chinese, we build the WenyanBENCH evaluation dataset and conduct extensive experiments for detailed analysis. The main contributions are as follows:

\begin{itemize}
	\item We propose WenyanGPT, a large language model focused on Classical Chinese. It demonstrates superior performance and wide applicability in tasks such as punctuation, part-of-speech tagging, translation, etc.
	\item We release pre-training and instruction fine-tuning datasets, along with a novel method for constructing domain-specific fine-tuning data, providing valuable resources for future research.
	\item We introduce WenyanBENCH, an evaluation benchmark for Classical Chinese tasks, with extensive experiments verifying WenyanGPT's leading performance across multiple tasks.
\end{itemize}

\section{Related Work}

\subsection{PLMs}

In 2017, Google introduced a new neural network architecture, Transformer. It utilizes self-attention mechanisms to better handle long-distance dependencies and significantly improves training efficiency through parallel computation. Based on Transformer, various LLMs have been proposed. BERT employs an encoder-only Transformer architecture and is pre-trained using masked language modeling and next sentence prediction tasks. The GPT series, on the other hand, uses a decoder-only Transformer architecture and an autoregressive language model (ALM). Over the course of the GPT series, the model size has steadily increased, from the original GPT to subsequent iterations, including GPT-2 \cite{Radford2019LanguageMA}, GPT-3 \cite{brown2020language}, and GPT-4 \cite{achiam2023gpt}, with continuous improvements in performance. PaLM \cite{chowdhery2023palm} uses the standard Transformer architecture in a decoder-only model with a modified SwiGLU activation function. This model, with 540 billion parameters, achieved human-level performance in 1-shot learning on the BIG-bench \cite{Srivastava2023BeyondTI} dataset. In 2023, Meta AI released the LLaMA model \cite{Touvron2023LLaMAOA}. This model also follows a decoder-only Transformer architecture and excels in various natural language processing tasks after large-scale training. In 2024, LLaMA 3 \cite{dubey2024llama} was released, including a pre-trained version with 405 billion parameters and a post-training version, alongside the LLaMA Guard 3 model for input-output safety. Pre-trained language models have seen rapid development, with Transformer-based models becoming the mainstream technology in natural language processing (NLP).

\subsection{Classical Chinese PLMs}
Pre-trained language models have achieved widespread success in the field of natural language processing. However, studies show that general-domain models often lack specialized knowledge for tasks in specific domains. Models pre-trained with domain-specific data tend to perform better for specialized tasks \cite{Ke2023ContinualPO,Gupta2023ContinualPO,Ibrahim2024SimpleAS,Taylor2022GalacticaAL,Lehman2023DoWS,Liu2020FinBERTAP}. In the field of Classical Chinese, several studies have extended models such as BERT, RoBERTa, and GPT by incorporating Classical Chinese corpora for pre-training, resulting in specialized models like AnchiBERT \cite{Tian2020AnchiBERTAP}, SikuBERT and SikuRoBERTa \cite{TSGL202206005}, GujiBERT and GujiGPT series \cite{wang2023gujibert}, and SikuGPT \cite{Liu2023SikuGPTAG}. These models show improved performance over general pre-trained models in Classical Chinese tasks. Instruction fine-tuning is another effective strategy. Using Supervised Fine-Tuning (SFT) can activate LLMs' ability to understand and answer questions in a specific domain \cite{Liu2023ChatCounselorAL,Xiong2023DoctorGLMFY,Wang2023HuaTuoTL,Yue2023DISCLawLLMFL,Huang2023LawyerLT,cui2023chatlaw,yang2023investlmlargelanguagemodel,Zhang2023XuanYuan2A,dan2023educhatlargescalelanguagemodelbased}. Classical Chinese LLMs \cite{TSGL202410011,yang2024zhongjing,Cao2023TranslatingAC,Cao2024TongGuMC} are in the early stages of development. For example, the "Xunzi\footnote{\url{ https://github.com/ Xunzi-LLM-of-Chinese-classics/ XunziALLM.}}" Classical Chinese large language model was trained on Classical Chinese-related corpora based on general models such as Qwen2.5 \cite{yang2024qwen2}, Baichuan2 \cite{yang2023baichuan}, and GLM-4 \cite{Zeng2024ChatGLMAF}. It has shown excellent performance in intelligent tagging, information extraction, and other tasks. TongGu \cite{Cao2024TongGuMC}, through two-stage instruction fine-tuning, is capable of Classical Chinese punctuation, translation, and appreciation tasks. In this paper, WenyanGPT is fine-tuned on higher-quality pre-training data and a larger, more diverse instruction dataset, showing superior and more comprehensive task handling capabilities.

\begin{figure}
	\centering
	\includegraphics[width=0.48\textwidth]{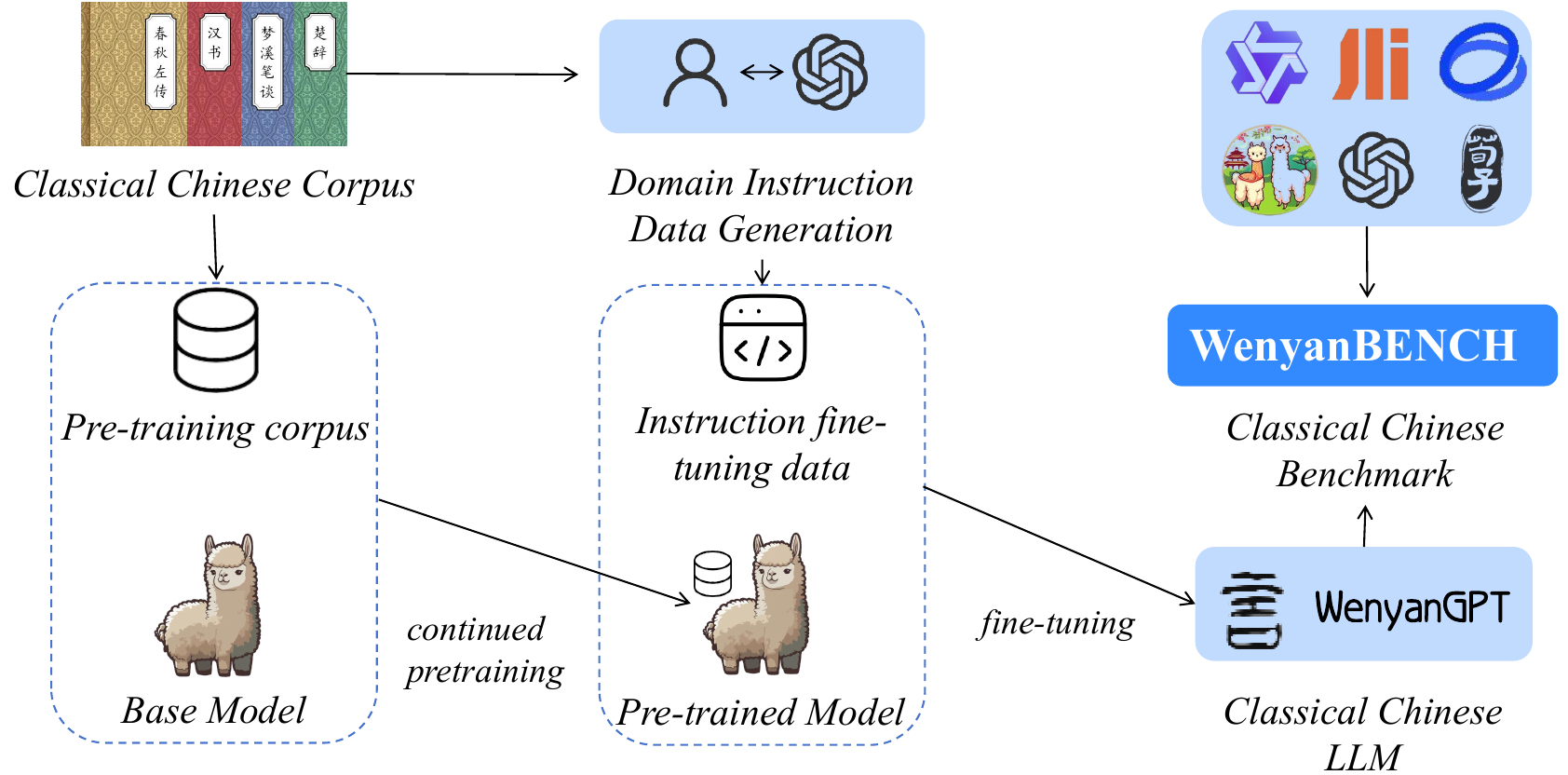}
	\caption{Overall Training Framework of WenyanGPT.}
	\label{fig:2}
\end{figure}

\section{WenyanGPT} 
In order to obtain the WenyanGPT Classical Chinese model, we first construct a Classical Chinese pre-training corpus and continue pre-training based on LLaMA3-8B-Chinese (Section 3.1). Then, we propose a method to construct domain instruction data (Section \ref{sft}). In our framework, instruction generation is manually constructed, guided by LLMs, and tested to ensure the high quality of the fine-tuning data. The complete training process is shown in Figure \ref{fig:2}.

\begin{table}
	\centering
	\setlength{\tabcolsep}{1mm}
	\begin{tabular}{cccc}
		\hline
		\textbf{Source} & \textbf{Scale} & \textbf{Source} & \textbf{Scale} \\
		\hline
		Daizhige & 5.2G & Poetry-master & 323M \\
		wenyanguji.com & 1.6G & PoetrySplider & 16M \\
		network resource & 1.1G & poems-db & 660M \\
		TCM-Ancient-Books & 322M & core-texts & 232M\\
		chinese-novel & 294M & sidamingzhi & 6.7M \\
		chinese-gushiwen & 23M & chtxt-main & 88M \\
		Classical-Chinese & 208M & chinese-poetry & 115M \\
		Classical-Modern & 853M & guner2023 & 63M \\
		core-books-main & 752M & kangxi-master & 37M \\
		GuWen-master & 2.5M & scripta-sinica & 3.7G \\
		\hline
	\end{tabular}
	\caption{Sources and Scale of Classical Chinese Pre-training Corpus.}
	\label{tab:1}
\end{table}

\subsection{Pre-training}

\begin{table}[h]
	\centering
	\setlength{\tabcolsep}{1mm}
	\begin{tabular}{cc}
		\hline
		\textbf{Hyper parameter}	&\textbf{Value} \\
		\hline
		per device train batch size	&16 \\
		gradient accumulation steps	&1 \\
		learning rate	&1.0e-4 \\
		num train epochs	&1 \\
		lr scheduler type	&cosine \\
		warmup ratio	&0.1\\
		\hline
	\end{tabular}
	\caption{Hyper-parameter Settings in Continue Pre-training.}
	\label{tab:2}
\end{table}

The corpus used in the pre-training phase is sourced from authoritative websites such as Daizhige, Wenyanguji, and various Classical Chinese-related data collected and organized from GitHub. The detailed data sources and scale are shown in Table \ref{tab:1}. We uniformly format and store data from these different sources, removing redundant information, errors, special symbols, and invalid characters. As a result, we obtain a clean, large-scale, high-quality Classical Chinese corpus of approximately 16GB. This corpus covers the Four Books and Five Classics, including Confucian classics, historical records, works of various philosophers, poetry, essays, drama, novels, miscellanies, and other literary genres. It also encompasses diverse fields such as local gazetteers, genealogies, religious texts, agricultural, law, medicine, astronomy, geography, craft books, and military texts. The corpus integrates both simplified and traditional Chinese characters, spanning from the Pre-Qin period to the Republic of China, providing rich materials for deep learning and research on ancient Chinese texts. We select LLaMA3-8B-Chinese as the base model and use the bfloat16 data format during training to improve efficiency. The hyper-parameter settings in pre-training are shown in Table \ref{tab:2}.

\begin{figure*}
	\centering
	\includegraphics[width=1\textwidth]{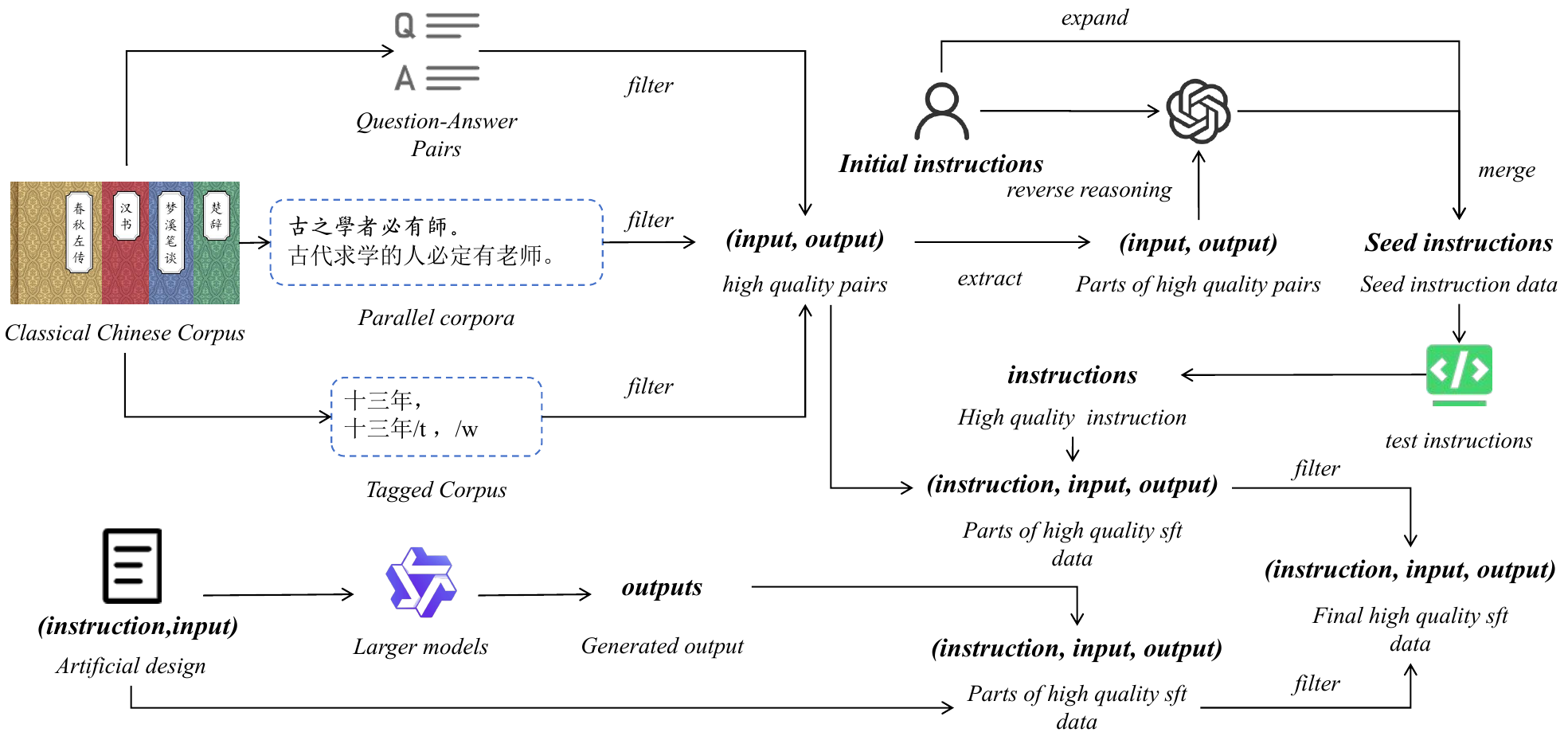}
	\caption{Instruction Fine-Tuning Data Construction Process.}
	\label{fig:3}
\end{figure*}

\subsection{Supervised Fine-Tuning}
\label{sft}
Based on continued pre-training, we perform supervised fine-tuning to better adapt the model to specific tasks and instructions. We use a high-quality set of instruction fine-tuning data, which we have previously collected and organized, to trigger the knowledge the model acquires during pre-training. The detailed process for constructing the instruction fine-tuning data is shown in Figure \ref{fig:3}. 

\paragraph{Data Selection and Initial Organization.}We select relevant data from the Classical Chinese corpus, including three major categories: question-answer pairs, parallel corpora, and tagged corpora. Parallel corpora are used for translation and interpretation tasks, while tagged corpora support fine-grained tasks such as punctuation and part-of-speech tagging.
In cases where the corpus lacks clear question-answer pairs, we generate supplementary data. During the selection phase, we prioritize high-quality data with clear content, standardized semantics, and task relevance to build the initial high-quality input-output pairs. 

\paragraph{Manual Design of Task Instructions and Model Expansion.}We design initial task instruction templates manually based on high-quality input and output, covering tasks such as Classical Chinese punctuation, and translating Classical Chinese into Modern Chinese. Then, we use LLMs, such as GPT and Qwen series to expand the task instructions. On the one hand, we generate diverse instructions from existing ones; on the other hand, we allow LLMs to perform reverse reasoning from the existing high-quality input-output pairs to generate new instructions, ensuring diversity in the instructions. After the expansion, we conduct an initial screening of the generated instructions, removing those with unclear or unreasonable semantics, resulting in a seed instruction set.

\paragraph{Testing the Instruction Set and Optimizing Fine-Tuning Data.}We randomly select high-quality input-output pairs and combine them with the seed instruction set to evaluate the model's adherence to instructions across different task scenarios. We analyze the test results to refine the instructional design and identify the task instructions that produce higher-quality outputs. Then, we combine the optimized instruction set with high-quality input-output pairs to construct fine-tuning data for specific instructions, providing reliable data support for subsequent model training.

\paragraph{Generation and Supplementation of Specific Task Data.} Due to the lack of specific task instruction data in the corpus, we manually design diverse initial instructions and inputs. Then we use LLMs including Qwen2.5-14B and Qwen2.5-72B to generate high-quality answers. We remove low-quality or irrelevant content through manual selection and automatic quality checks, forming another portion of the instruction data.

\begin{table}[t]
	\centering
	\setlength{\tabcolsep}{1pt}
	\begin{tabular}{ccc}
		\hline
		\textbf{Task}	&\textbf{Data Source}	&\textbf{Num} \\
		\hline
		Punctuation	&Daizhige	&107,3017 \\
		Part-of-speech tagging	&evahan	&9,952\\
		NER	&Self-built	&29923\\
		Translation	&classical-modern	&222,700\\
		&wenyanguji.com	    &302,724\\
		Word explanation	&gushiwen.com	&31,088\\
		Reverse dictionary	&chinese-dictionary	&39,708\\
		&chinese-xinhua	&138,810\\
		Total		&       &1,847,922\\
		\hline
	\end{tabular}
	\caption{Sources and Scale of Instruction Fine-Tuning Data.}
	\label{tab:3}
\end{table}

\begin{table}[t]
	\centering
	\setlength{\tabcolsep}{1mm}
	\begin{tabular}{cc}
		\hline
		\textbf{Hyper parameter}	&\textbf{Value} \\
		\hline
		per device train batch size	&8 \\
		gradient accumulation steps	&2 \\
		learning rate	&1.0e-4 \\
		num train epochs	&1 \\
		lr scheduler type	&cosine \\
		warmup ratio	&0.1\\
		\hline
	\end{tabular}
	\caption{Hyper-parameter Settings in Fine-Tuning.}
	\label{tab:4}
\end{table}

\paragraph{Data Integration and Output Validation.} We integrate the data generated from the corpus and the supplementary data from the LLMs to form a complete instruction dataset. During integration, we ensure consistency between instructions and outputs, covering multiple Classical Chinese task scenarios. Finally, through comprehensive data validation and optimization, we create a high-quality instruction dataset. In the end, we obtain about 1.85 million instruction fine-tuning data. The detailed data sources and counts are shown in Table \ref{tab:3}. We use these data to fine-tune the pre-trained model. The hyper-parameter settings in fine-tuning are shown in Table \ref{tab:4}.

\section{Benchmarking Classical Chinese Tasks}

\begin{table}
	\centering
	\setlength{\tabcolsep}{3mm}
	\begin{tabular}{cc}
		\hline
		\textbf{Task} 	&\textbf{Num}  \\
		\hline
		Punctuation		&7,559  \\
		Part-of-speech tagging	&1,247  \\
		NER	 	&3,741   \\
		Translation	&5,013  \\
		Word Explanation	&3,931  \\
		Reverse Dictionary	&4,462  \\
		Total	   &25,953\\
		\hline
	\end{tabular}
	\caption{Data Sources and Detailed Statistics of WenyanBench.}
	\label{tab:5}
\end{table}

\begin{table*}
	\centering
	\setlength{\tabcolsep}{4pt}
	\renewcommand{\arraystretch}{}
	\begin{tabular*}{0.98\linewidth}{cccccccccc}
		\hline
		\textbf{Model} & \multicolumn{3}{c}{\textbf{Punctuation}} & \multicolumn{3}{c}{\textbf{Part-of-speech tagging}} & \multicolumn{3}{c}{\textbf{NER}}\\
		& \textbf{P(\%)} & \textbf{R(\%)} & \textbf{F1(\%)} & \textbf{P(\%)} & \textbf{R(\%)} & \textbf{F1(\%)} & \textbf{P(\%)} & \textbf{R(\%)} & \textbf{F1(\%)} \\
		\hline
		Qwen2.5-7B-Instruct	&54.34	&53.31	&53.82	&51.25	&48.16	&49.65	&66.05	&46.55	&54.61 \\
		Baichuan2-7B-Chat	&51.05	&21.03	&29.79	&47.11	&30.97	&37.37	&35.10	&10.58	&16.26 \\
		GLM-4-9B-Chat	&52.39	&55.00	&53.66	&49.90	&54.98	&52.32	&52.19	&45.42 &48.57\\
		Meta-Llama-3-8B-Instruct	&55.05	&22.41	&31.85	&25.73	&17.06	&20.52	&47.50	&57.48	&52.01\\
		Llama3-8B-Chinese-Chat	&45.76	&38.07	&41.56	&21.34	&19.34	&20.29	&46.85	&66.69	&55.04\\
		Xunzi-Qwen-1.5-7B-Chat	&52.08	&47.19	&49.51	&77.54	&78.07	&77.81	&49.79	&51.21	&50.49\\
		GPT-4o	&52.00	&50.70	&51.34	&82.41	&81.11	&\underline{81.75}	&61.58	&76.97	&\underline{68.42}\\
		Deepseek-V3 &56.33  &61.94  &\underline{59.01}  &79.12  &79.18  &79.15  &56.83  &79.75  &66.36 \\
		\textbf{WenyanGPT}	&\textbf{76.84}	&\textbf{74.52}	&\textbf{75.66}	&\textbf{89.66}	&\textbf{88.54}	&\textbf{89.1}	&\textbf{92.14}	&\textbf{90.19}	&\textbf{91.16}\\
		\hline
	\end{tabular*}
	\caption{Results for understanding tasks (Punctuation, Part-of-speech tagging, Named Entity Recognition) on WenyanBench. The results underlined represent the second-best model's F1 score.}
	\label{tab:7}
\end{table*}

\begin{figure*}[t]
	\centering
	\includegraphics[width=1\textwidth]{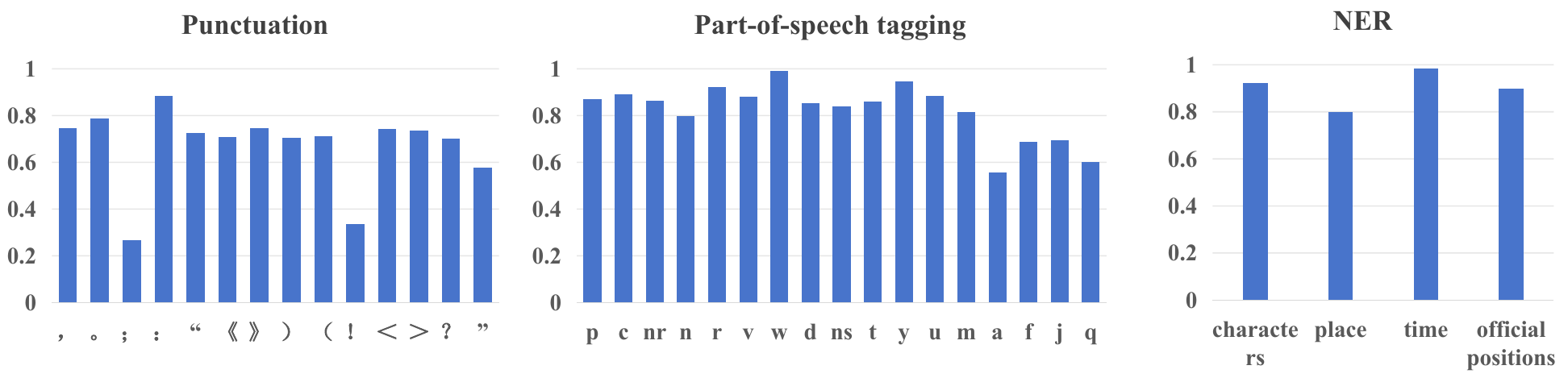}
	\caption{The F1 scores of WenYanGPT for the subcategories of understanding tasks (including Punctuation, Part-of-speech tagging, and Named Entity Recognition) on WenyanBench.}
	\label{fig:4}
\end{figure*}

\paragraph{WenyanBench.} In order to evaluate the model's performance on Classical Chinese tasks, we devise a benchmark known as WenyanBench. WenyanBench shares the same data sources as the instruction fine-tuning data and has undergone duplicate data removal, as well as validation by both manual and LLMs. For quality control, we sample a subset of the data. The distribution and detailed statistics of WenyanBench are shown in Table \ref{tab:5}.

\paragraph{Tasks.} Our benchmark includes six tasks related to Classical Chinese. Among them, we subdivide 14 types of punctuation marks in the punctuation task, divide ancient Chinese word classes into 17 categories in the part-of-speech tagging task, and define 4 categories for the named entity recognition task.

\paragraph{Metrics.} For the WenyanBench benchmark, different evaluation metrics are used for different types of tasks (understanding tasks and generation tasks). For understanding tasks, evaluation primarily relies on Precision, Recall, and F1-Score. For generation tasks, BLEU and BERT-Score are used as evaluation metrics. BLEU measures the N-gram overlap between generated content and reference answers, while BERT-Score better captures the semantic similarity between the generated content and reference answers.

\paragraph{Evaluation Method.} To efficiently assess model performance, we design a set of scripted tools to automatically compute BLEU, BERT-Score, and other metrics. These tools quickly and accurately quantify model outputs, providing clear feedback for model optimization. This automated evaluation approach improves evaluation efficiency and ensures the consistency and comparability of the results.

\section{Experiments}

The experiments evaluate WenyanGPT's performance on understanding and generation tasks in Classical Chinese.

\subsection{Experimental Setup}
\paragraph{Baselines.} The baselines include general-domain and Classical Chinese domain LLMs. The general-domainLLMs are Qwen2.5-7B-Instruct, Baichuan2-7B-Chat, GLM-4-9B-Chat, Meta-Llama-3-8B-Instruct, Llama3-8B-Chinese-Chat, GPT-4o, and Deepseek-V3 \cite{deepseekai2025deepseekv3technicalreport}. The Classical Chinese domain LLM is Xunzi-Qwen1.5-7B-Chat.

\paragraph{Data and Evaluation.} We use the WenyanBench benchmark for testing. The understanding tasks, including punctuation, part-of-speech tagging, and named entity recognition, are evaluated by Precision, Recall, and F1-Score. The generation tasks include word explanation, translation, and reverse dictionary, where BLEU is used for word explanation and translation, and BERT-Score is used for reverse dictionary.

\begin{table*}
	\centering
	\setlength{\tabcolsep}{2pt}
	\renewcommand{\arraystretch}{}{
		\begin{tabular}{cccccccccccc}
			\hline
			\textbf{Model} & \multicolumn{4}{c}{\textbf{Translation}} & \multicolumn{4}{c}{\textbf{Word explanation}} & \multicolumn{3}{c}{\textbf{Reverse dictionary}} \\
			& \textbf{Bleu1} & \textbf{Bleu2} & \textbf{Bleu3} & \textbf{Bleu4} & \textbf{Bleu1} & \textbf{Bleu2} & \textbf{Bleu3} & \textbf{Bleu4} & \textbf{P(\%)} & \textbf{R(\%)} & \textbf{F1(\%)} \\
			
			\hline
			Qwen2.5-7B-Instruct	&0.37 	&0.23 	&0.17 	&0.14 	&0.16 	&0.09 	&0.07 	&0.05 	&68.43	&68.99	&68.66\\
			Baichuan2-7B-Chat	&0.33 	&0.20 	&0.14 	&0.11 	&0.14 	&0.08 	&0.05 	&0.04 	&64.81	&66.21	&65.42 \\
			GLM-4-9B-Chat	&0.34 	&0.21 	&0.15 	&0.12 	&0.15 	&0.09 	&0.06 	&0.05 	&65.58	&68.04	&66.69\\
			Meta-Llama-3-8B-Instruct	&0.16 	&0.09 	&0.06 	&0.05 	&0.11 	&0.06 	&0.05 	&0.04 	&59.41	&64.13	&61.48\\
			Llama3-8B-Chinese-Chat	&0.26 	&0.15 	&0.10 	&0.08 	&0.11 	&0.06 	&0.04 	&0.03 	&61.8	&65.18	&63.28\\
			Xunzi-Qwen1.5-7B-Chat	&0.22 	&0.15 	&0.11 	&0.09 	&0.11 	&0.08 	&0.06 	&0.05 	&66.47	&68.45	&67.35\\
			GPT-4o	&\underline{0.41} 	&0.27 	&0.19 	&0.14   &0.19	&0.13 	&0.09 	&0.07 	&64.96	&66.76	&65.81\\
			Deepseek-V3 &0.30	&0.19	&0.13	&0.10   &\underline{0.20} 	&0.14 	&0.11 	&0.08	&71.93	&71.84	&\underline{71.88} \\
			\textbf{WenyanGPT}	&\textbf{0.47} 	&\textbf{0.33} 	&\textbf{0.24} 	&\textbf{0.19} 	&\textbf{0.35}	&\textbf{0.31}	&\textbf{0.27}	&\textbf{0.23}	&\textbf{75.51}	&\textbf{75.31}	&\textbf{75.39}\\
			\hline
		\end{tabular}
	}
	\caption{Results for generation tasks (Translation, Word explanation, Reverse dictionary) on WenyanBench. The results underlined represent the second-best model's BLEU1 score and BERT-Score-F1 score.}
	\label{tab:8}
\end{table*}

\begin{figure}[t]
	\centering
	\includegraphics[width=0.5\textwidth]{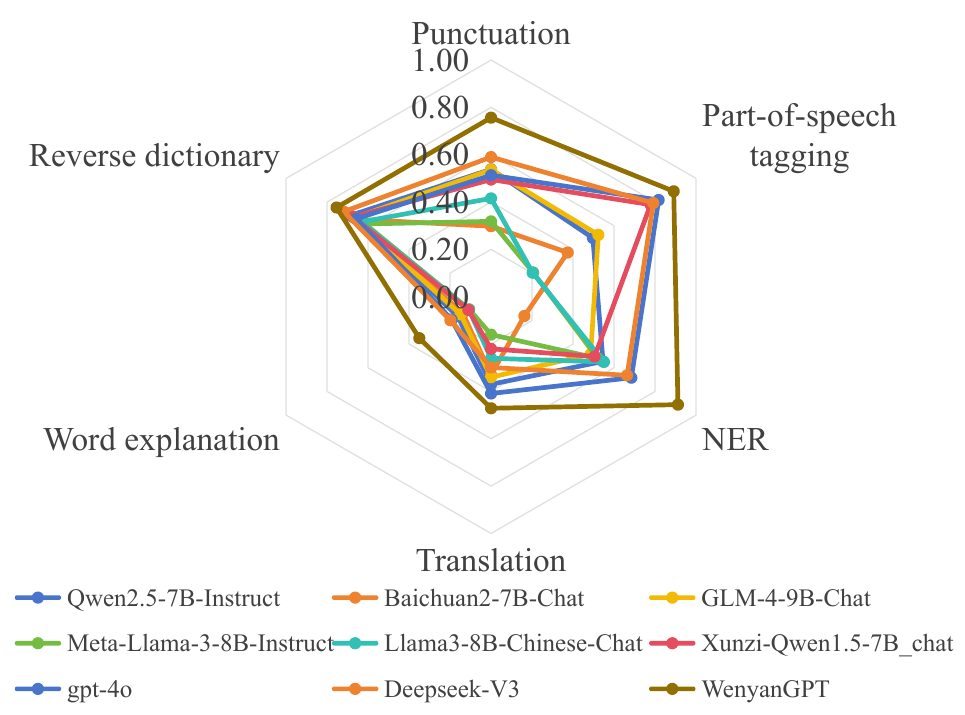}
	\caption{Radar plot showing model performance on WenyanBench, with values normalized to a 0-1 scale.}
	\label{fig:5}
\end{figure}

\subsection{Experimental Analysis}

\paragraph{WenyanGPT demonstrates a significant lead in language understanding tasks.} The experimental results for understanding tasks are presented in Table \ref{tab:7}. In the named entity recognition task, WenyanGPT's precision, recall, and F1 score all exceed 90\%, while the second-best model, GPT-4o, fails to surpass 80\% in any of these metrics. In the punctuation task, WenyanGPT's F1 score is 16.65\% higher than that of the second-best model, Deepseek-V3, reaching 75.66\%. Additionally, in the part-of-speech tagging task, WenyanGPT's F1 score is 7.35\% higher than that of the second-best model, GPT-4o. These results highlight WenyanGPT's overwhelming superiority in understanding tasks, particularly in NER and punctuation. This performance is attributed to the model's extensive pre-training on Classical Chinese data, enabling it to handle complex linguistic phenomena and ensuring higher accuracy and stability in fundamental language understanding tasks, such as part-of-speech tagging and NER.

\paragraph{WenyanGPT demonstrates excellent capabilities and high efficiency in subcategories of Classical Chinese comprehension tasks.} As shown in Figure \ref{fig:4}, the F1 scores of our model in the three task's subcategories are generally stable and high. Specifically, in the named entity recognition task, WenyanGPT's F1 score remains above 80\%, demonstrating its strong ability to correctly identify entities such as historical figures, place, and proper nouns in ancient Chinese texts. This performance shows that it has strong accuracy and robustness when processing ancient Chinese texts, and can effectively capture complex contextual relationships and word meaning changes. Overall, WenyanGPT's high F1 score in ancient Chinese comprehension tasks not only reflects its high efficiency in basic tasks, but also demonstrates its advantages and potential in processing fine-grained tasks in ancient Chinese.

\paragraph{WenyanGPT can significantly enhance the quality of generated content in generation tasks.} The experimental results for generation tasks are displayed in Table \ref{tab:8}. WenyanGPT exhibits strong semantic preservation and contextual consistency, demonstrates superior and more stable performance in translation and word explanation tasks, with consistently high BLEU scores (BLEU1-BLEU4). In classical Chinese translation, the model's BLEU1 score of 0.47 exceeds that of the second-best model by 0.06. Furthermore, WenyanGPT's F1 score in the reverse dictionary task is 3.47\% higher than the second best model. Through meticulous pre-training and multitask training, WenyanGPT develops robust contextual consistency, enabling it to generate content that accurately reflects the intended meaning and context. This capability ensures high-quality semantic transmission and content coherence in both long-text generation and complex tasks, such as word explanation.

\paragraph{WenyanGPT can achieve better performance than the previous open source LLMs in Classical Chinese tasks.} WenyanGPT, through large-scale pre-training and multitask joint optimization, demonstrates substantial capabilities in Classical Chinese processing tasks, establishing a clear lead over existing mainstream LLMs. This validates the effectiveness of the proposed approach. The performance of various LLMs on WenyanBench is shown in Figure \ref{fig:5}. It is evident that WenyanGPT has achieved the highest score in six Classical Chinese tasks. WenyanGPT's multitask training strategy, which integrates various Classical Chinese processing tasks, enhances the model's ability to learn across tasks. The mutual reinforcement of shared linguistic features and semantic information between tasks notably boosts the model's generalization capabilities. This inter-task synergy not only improves performance on individual tasks but also enables WenyanGPT to effectively handle multiple tasks simultaneously, maintaining stable performance, especially in complex tasks, and demonstrating superior adaptability across tasks.

\begin{table}
	\centering
	\renewcommand{\arraystretch}{}
	\begin{tabular}{p{2.5cm}p{4.6cm}}
		\hline
		\textbf{Case 1}
		& Please segment the following Classical Chinese content and accurately tag the parts of speech: \begin{CJK}{UTF8}{gbsn}四年春，衞州吁弑桓公而立。\end{CJK}  \\
		\hline
		\textbf{Ground Truth} 
		& \begin{CJK}{UTF8}{gbsn}四年/t 春/n ，/w 衞州吁/nr 弑/v 桓公/nr 而/c 立/v 。/w\end{CJK}  \\
		\hline
		\textbf{WenyanGPT}
		& \begin{CJK}{UTF8}{gbsn}四年/t 春/n ，/w 衞州吁/nr 弑/v 桓公/nr 而/c 立/v 。/w\end{CJK} \\
		Deepseek-V3
		& \begin{CJK}{UTF8}{gbsn}{\textcolor{red}{四/m 年/t 春/t}}，/w {\textcolor{red}{衞/ns}} 州吁/nr 弑/v 桓公/nr 而/c 立/v 。/w\end{CJK} \\
		GPT-4o
		& \begin{CJK}{UTF8}{gbsn}{\textcolor{red}{四年/n}} 春/n ，/w {\textcolor{red}{衞州/n}} {\textcolor{red}{吁/v}} 弑/v 桓公/n 而/c 立/v 。/w\end{CJK} \\
		Qwen2.5-7B-Instruct
		& \begin{CJK}{UTF8}{gbsn}四年/t {\textcolor{red}{春/w}} ，/w {\textcolor{blue}{卫}}州吁/nr {\textcolor{blue}{射}}/v 桓公/nr 而/c 立/v 。/w\end{CJK} \\
		Xunzi-Qwen1.5-7B-Chat
		& \begin{CJK}{UTF8}{gbsn}{\textcolor{red}{四年春/t}} ，/w 衞州吁/nr {\textcolor{blue}{杀}}/v 桓公/nr 而/c 立/v 。/w\end{CJK} \\
		\hline
	\end{tabular}%
	\caption{Response examples for the part-of-speech tagging task from different LLMs. Errors in part-of-speech tagging are marked in red, while text errors are highlighted in blue.}
	\label{tab:9}%
\end{table}%

\subsection{Case Study}
We provide response examples for part-of-speech tagging from five LLMs: WenyanGPT, Deepseek-V3, GPT-4o \cite{hurst2024gpt}, Qwen2.5-7B-Instruct, and Xunzi-Qwen1.5-7B-Chat, as shown in Table \ref{tab:9}. By analyzing these examples, we can see the performance differences of different LLMs in ccessing tasks, especially in the understanding tasks of classical Chinese.

Table \ref{tab:9} illustrates some typical errors in the \textbf{part-of-speech tagging} task. Specifically, GPT-4o makes mistakes in tagging time words and proper nouns. For instance, it incorrectly tags the time expression "\begin{CJK}{UTF8}{gbsn}四年\end{CJK}" (the forth year) as a common noun, and it also struggles with recognizing the proper noun "\begin{CJK}{UTF8}{gbsn}衞州吁\end{CJK}" (Wei Zhou Xu). Qwen2.5-7B-Instruct errors in tagging the part of speech for "\begin{CJK}{UTF8}{gbsn}春\end{CJK}" (spring) and sometimes replaces characters in the original text with inappropriate alternatives. Xunzi-Qwen1.5-7B-Chat primarily fails in distinguishing between time words and nouns, mistakenly using the simplified character "\begin{CJK}{UTF8}{gbsn}杀\end{CJK}" instead of the correct "\begin{CJK}{UTF8}{gbsn}弑\end{CJK}". These errors highlight the model's difficulty in making precise distinctions between similar words. 

WenyanGPT demonstrates strong semantic understanding and retention capabilities in handling Classical Chinese tasks. Its ability to accurately tag parts of speech and generate concise, poetic word explanations demonstrates a deep grasp of the nuances of Classical Chinese. In contrast, other LLMs struggle with both understanding the subtle distinctions of Classical Chinese and producing responses that remain faithful to the content and meaning of the original text. This positions WenyanGPT as a powerful tool for handling complex tasks in the domain of Classical Chinese, far surpassing the other LLMs in terms of both accuracy and literariness.

\section{Conclusion}

We propose a comprehensive solution for Classical Chinese language processing challenges, including the development of WenyanGPT, a large language model focused on the Classical Chinese domain, and WenyanBENCH, an evaluation benchmark dataset for Classical Chinese tasks. We release pre-training and instruction fine-tuning datasets and describe the method for constructing the instruction fine-tuning dataset. Through systematic experiments and analysis, we demonstrate the significant impact of domain-specific pre-training and multi-task instruction fine-tuning on improving Classical Chinese processing capabilities. Our model outpermforms existing mainstream LLMs in various downstream tasks. In future, we attend to explore the potential of multimodal models by combining Classical Chinese texts with image data (such as inscriptions and manuscripts) to enhance processing capabilities.

\section*{Limitations}

Although WenyanGPT has made progress in Classical Chinese tasks, there are still some limitations. Firstly, due to space limitations and the subjectivity of evaluation (manual assessment), tasks such as poetry generation are not included in the paper. Secondly, the model mainly relies on large-scale, high-quality instruction datasets. Finally, there is still room for improvement when handling long Classical Chinese texts and complex syntax.

\bibliography{WenyanGPT}

\begin{thebibliography}{50}
\providecommand{\natexlab}[1]{#1}

\bibitem[{Achiam et~al.(2023)Achiam, Adler, Agarwal, Ahmad, Akkaya, Aleman, Almeida et~al.}]{achiam2023gpt}
Josh Achiam, Steven Adler, Sandhini Agarwal, Lama Ahmad, Ilge Akkaya, Florencia~Leoni Aleman, Diogo Almeida, and 1 others. 2023.
\newblock Gpt-4 technical report.
\newblock \emph{arXiv preprint arXiv:2303.08774}.

\bibitem[{Brown et~al.(2020)Brown, Mann, Ryder, Subbiah et~al.}]{brown2020language}
Tom Brown, Benjamin Mann, Nick Ryder, Melanie Subbiah, and 1 others. 2020.
\newblock Language models are few-shot learners.
\newblock \emph{Advances in neural information processing systems}, 33:1877--1901.

\bibitem[{Cao et~al.(2023)Cao, Peng, Shi, Jiang, and Jin}]{Cao2023TranslatingAC}
Jiahuan Cao, Dezhi Peng, Yongxin Shi, Zongyuan Jiang, and Lianwen Jin. 2023.
\newblock \href {https://api.semanticscholar.org/CorpusID:265158334} {Translating ancient chinese to modern chinese at scale: A large language model-based approach}.
\newblock In \emph{International Conference on Algorithmic Learning Theory}.

\bibitem[{Cao et~al.(2024)Cao, Peng, Zhang, Shi et~al.}]{Cao2024TongGuMC}
Jiahuan Cao, Dezhi Peng, Peirong Zhang, Yongxin Shi, and 1 others. 2024.
\newblock \href {https://api.semanticscholar.org/CorpusID:271039137} {Tonggu: Mastering classical chinese understanding with knowledge-grounded large language models}.
\newblock In \emph{Conference on Empirical Methods in Natural Language Processing}.

\bibitem[{Chang et~al.(2024)Chang, Yuan, Li, Xu et~al.}]{XDTQ202411008}
Bolin Chang, Yiguo Yuan, Bin Li, Zhixing Xu, and 1 others. 2024.
\newblock Automatic word segmentation and part-of-speech tagging for classical chinese based on radicals.
\newblock \emph{Data Analysis and Knowledge Discovery}, 8(11):102--113.

\bibitem[{Cheng et~al.(2020)Cheng, Li, Xiao, Xu et~al.}]{cheng2020integration}
Ning Cheng, Bin Li, Liming Xiao, Changwei Xu, and 1 others. 2020.
\newblock Integration of automatic sentence segmentation and lexical analysis of ancient chinese based on bilstm-crf model.
\newblock In \emph{Proceedings of LT4HALA 2020-1st Workshop on Language Technologies for Historical and Ancient Languages}, pages 52--58.

\bibitem[{Chowdhery et~al.(2023)Chowdhery, Narang, Devlin, Bosma et~al.}]{chowdhery2023palm}
Aakanksha Chowdhery, Sharan Narang, Jacob Devlin, Maarten Bosma, and 1 others. 2023.
\newblock Palm: Scaling language modeling with pathways.
\newblock \emph{Journal of Machine Learning Research}, 24(240):1--113.

\bibitem[{Cui et~al.(2023)Cui, Li, Yan, Chen, and Yuan}]{cui2023chatlaw}
Jiaxi Cui, Zongjian Li, Yang Yan, Bohua Chen, and Li~Yuan. 2023.
\newblock Chatlaw: Open-source legal large language model with integrated external knowledge bases.
\newblock \emph{CoRR}.

\bibitem[{Dan et~al.(2023)Dan, Lei, Gu, Li et~al.}]{dan2023educhatlargescalelanguagemodelbased}
Yuhao Dan, Zhikai Lei, Yiyang Gu, Yong Li, and 1 others. 2023.
\newblock \href {https://arxiv.org/abs/2308.02773} {Educhat: A large-scale language model-based chatbot system for intelligent education}.
\newblock \emph{Preprint}, arXiv:2308.02773.

\bibitem[{DeepSeek-AI et~al.(2025)DeepSeek-AI, Liu, Feng, Xue, Wang, Wu, Lu, Zhao, Deng, Zhang, Ruan, Dai, Guo, Yang, Chen, Ji, Li, Lin, Dai, Luo, Hao, Chen, Li, Zhang, Bao, Xu, Wang, Zhang, Ding, Xin, Gao, Li, Qu, Cai, Liang, Guo, Ni, Li, Wang, Chen, Chen, Yuan, Qiu, Li, Song, Dong, Hu, Gao, Guan, Huang, Yu, Wang, Zhang, Xu, Xia, Zhao, Wang, Zhang, Li, Wang, Zhang, Zhang, Tang, Li, Tian, Huang, Wang, Zhang, Wang, Zhu, Chen, Du, Chen, Jin, Ge, Zhang, Pan, Wang, Xu, Zhang, Chen, Li, Lu, Zhou, Chen, Wu, Ye, Ye, Ma, Wang, Zhou, Yu, Zhou, Pan, Wang, Yun, Pei, Sun, Xiao, Zeng, Zhao, An, Liu, Liang, Gao, Yu, Zhang, Li, Jin, Wang, Bi, Liu, Wang, Shen, Chen, Zhang, Chen, Nie, Sun, Wang, Cheng, Liu, Xie, Liu, Yu, Song, Shan, Zhou, Yang, Li, Su, Lin, Li, Wang, Wei, Zhu, Zhang, Xu, Xu, Huang, Li, Zhao, Sun, Li, Wang, Yu, Zheng, Zhang, Shi, Xiong, He, Tang, Piao, Wang, Tan, Ma, Liu, Guo, Wu, Ou, Zhu, Wang, Gong, Zou, He, Zha, Xiong, Ma, Yan, Luo, You, Liu, Zhou, Wu, Ren, Ren, Sha, Fu, Xu, Huang, Zhang, Xie, Zhang, Hao,
  Gou, Ma, Yan, Shao, Xu, Wu, Zhang, Li, Gu, Zhu, Liu, Li, Xie, Song, Gao, and Pan}]{deepseekai2025deepseekv3technicalreport}
DeepSeek-AI, Aixin Liu, Bei Feng, Bing Xue, Bingxuan Wang, Bochao Wu, Chengda Lu, Chenggang Zhao, Chengqi Deng, Chenyu Zhang, Chong Ruan, Damai Dai, Daya Guo, Dejian Yang, Deli Chen, Dongjie Ji, Erhang Li, Fangyun Lin, Fucong Dai, and 181 others. 2025.
\newblock \href {https://arxiv.org/abs/2412.19437} {Deepseek-v3 technical report}.
\newblock \emph{Preprint}, arXiv:2412.19437.

\bibitem[{Dubey et~al.(2024)Dubey, Jauhri, Pandey et~al.}]{dubey2024llama}
Abhimanyu Dubey, Abhinav Jauhri, Abhinav Pandey, and 1 others. 2024.
\newblock The llama 3 herd of models.
\newblock \emph{arXiv preprint arXiv:2407.21783}.

\bibitem[{Gupta et~al.(2023)Gupta, Th'erien, Ibrahim, Richter, Anthony, Belilovsky, Rish, and Lesort}]{Gupta2023ContinualPO}
Kshitij Gupta, Benjamin Th'erien, Adam Ibrahim, Mats~L. Richter, Quentin~G. Anthony, Eugene Belilovsky, Irina Rish, and Timoth{\'e}e Lesort. 2023.
\newblock \href {https://api.semanticscholar.org/CorpusID:260704601} {Continual pre-training of large language models: How to (re)warm your model?}
\newblock \emph{ArXiv}, abs/2308.04014.

\bibitem[{Huang et~al.(2020)Huang, Lu, Cheng, and Peng}]{huang2020generating}
Chuen-Min Huang, Kuo-Lin Lu, Yi-Ying Cheng, and Yu-Chen Peng. 2020.
\newblock Generating chinese classical poetry with quatrain generation model (qgm) using encoder-decoder lstm.
\newblock In \emph{2020 IEEE International Conference on Big Data (Big Data)}, pages 5700--5702. IEEE.

\bibitem[{Huang et~al.(2010)Huang, Sun, and Chen}]{DBLP:conf/acl-sighan/HuangSC10}
Hen{-}Hsen Huang, Chuen{-}Tsai Sun, and Hsin{-}Hsi Chen. 2010.
\newblock \href {https://aclanthology.org/W10-4103/} {Classical chinese sentence segmentation}.
\newblock In \emph{{CIPS-SIGHAN} Joint Conference on Chinese Language Processing, Beijing, China, August 28-29, 2010}.

\bibitem[{Huang et~al.(2002)Huang, Peng, Wang, and Wu}]{10.5555/647240.718633}
Liang Huang, Yinan Peng, Huan Wang, and Zhenyu Wu. 2002.
\newblock Statistical part-of-speech tagging for classical chinese.
\newblock In \emph{Proceedings of the 5th International Conference on Text, Speech and Dialogue}, TSD '02, page 115–122, Berlin, Heidelberg. Springer-Verlag.

\bibitem[{Huang et~al.(2023)Huang, Tao, An, Zhang, Jiang, Chen, Wu, and Feng}]{Huang2023LawyerLT}
Quzhe Huang, Mingxu Tao, Zhenwei An, Chen Zhang, Cong Jiang, Zhibin Chen, Zirui Wu, and Yansong Feng. 2023.
\newblock \href {https://api.semanticscholar.org/CorpusID:258865862} {Lawyer llama technical report}.

\bibitem[{Hurst et~al.(2024)Hurst, Lerer, Goucher, Perelman et~al.}]{hurst2024gpt}
Aaron Hurst, Adam Lerer, Adam~P Goucher, Adam Perelman, and 1 others. 2024.
\newblock Gpt-4o system card.
\newblock \emph{arXiv preprint arXiv:2410.21276}.

\bibitem[{Ibrahim et~al.(2024)Ibrahim, Th'erien, Gupta, Richter, Anthony, Lesort, Belilovsky, and Rish}]{Ibrahim2024SimpleAS}
Adam Ibrahim, Benjamin Th'erien, Kshitij Gupta, Mats~L. Richter, Quentin Anthony, Timoth{\'e}e Lesort, Eugene Belilovsky, and Irina Rish. 2024.
\newblock \href {https://api.semanticscholar.org/CorpusID:268379604} {Simple and scalable strategies to continually pre-train large language models}.
\newblock \emph{ArXiv}, abs/2403.08763.

\bibitem[{Ke et~al.(2023)Ke, Shao, Lin, Konishi, Kim, and Liu}]{Ke2023ContinualPO}
Zixuan Ke, Yijia Shao, Haowei Lin, Tatsuya Konishi, Gyuhak Kim, and Bin Liu. 2023.
\newblock \href {https://api.semanticscholar.org/CorpusID:258079422} {Continual pre-training of language models}.
\newblock In \emph{International Conference on Learning Representations}.

\bibitem[{Kenton and Toutanova(2019)}]{kenton2019bert}
Jacob Devlin Ming-Wei~Chang Kenton and Lee~Kristina Toutanova. 2019.
\newblock Bert: Pre-training of deep bidirectional transformers for language understanding.
\newblock In \emph{Proceedings of naacL-HLT}, volume~1, page~2. Minneapolis, Minnesota.

\bibitem[{Lehman et~al.(2023)Lehman, Hernandez, Mahajan, Wulff, Smith, Ziegler, Nadler, Szolovits, Johnson, and Alsentzer}]{Lehman2023DoWS}
Eric Lehman, Evan Hernandez, Diwakar Mahajan, Jonas Wulff, Micah~J Smith, Zachary Ziegler, Daniel Nadler, Peter Szolovits, Alistair Johnson, and Emily Alsentzer. 2023.
\newblock \href {https://api.semanticscholar.org/CorpusID:256900662} {Do we still need clinical language models?}
\newblock \emph{ArXiv}, abs/2302.08091.

\bibitem[{Li(2018)}]{li2018automatic}
N~Li. 2018.
\newblock Automatic extraction of alias in ancient local chronicles based on conditional random fields.
\newblock \emph{J. Chin. Inf. Process}, 32:41.

\bibitem[{Liu et~al.(2023{\natexlab{a}})Liu, Wang, Zhao, Hu, Wu, Lin, Shen, Li, Liu, Zhang, and Zhao}]{Liu2023SikuGPTAG}
Chang Liu, Dongbo Wang, Zhixiao Zhao, Die Hu, Mengcheng Wu, Litao Lin, Si~Shen, Bin Li, Jiangfeng Liu, Hai Zhang, and Lianzheng Zhao. 2023{\natexlab{a}}.
\newblock \href {https://api.semanticscholar.org/CorpusID:258179714} {Sikugpt: A generative pre-trained model for intelligent information processing of ancient texts from the perspective of digital humanities}.
\newblock \emph{ArXiv}, abs/2304.07778.

\bibitem[{Liu et~al.(2018)Liu, Lv, Yang, and Qu}]{Liu2018AncientModernCT}
Dayiheng Liu, Jiancheng Lv, Kexin Yang, and Qian Qu. 2018.
\newblock \href {https://api.semanticscholar.org/CorpusID:51972516} {Ancient–modern chinese translation with a new large training dataset}.
\newblock \emph{ACM Transactions on Asian and Low-Resource Language Information Processing (TALLIP)}, 19:1 -- 13.

\bibitem[{Liu et~al.(2023{\natexlab{b}})Liu, Li, Cao, Ren, Liao, and Wu}]{Liu2023ChatCounselorAL}
June~M. Liu, Donghao Li, He~Cao, Tianhe Ren, Zeyi Liao, and Jiamin Wu. 2023{\natexlab{b}}.
\newblock \href {https://api.semanticscholar.org/CorpusID:262943261} {Chatcounselor: A large language models for mental health support}.
\newblock \emph{ArXiv}, abs/2309.15461.

\bibitem[{Liu et~al.(2020)Liu, Huang, Huang, Li, and Zhao}]{Liu2020FinBERTAP}
Zhuang Liu, Degen Huang, Kaiyu Huang, Zhuang Li, and Jun Zhao. 2020.
\newblock \href {https://api.semanticscholar.org/CorpusID:265038146} {Finbert: A pre-trained financial language representation model for financial text mining}.
\newblock In \emph{International Joint Conference on Artificial Intelligence}.

\bibitem[{Radford and Narasimhan(2018)}]{Radford2018ImprovingLU}
Alec Radford and Karthik Narasimhan. 2018.
\newblock \href {https://api.semanticscholar.org/CorpusID:49313245} {Improving language understanding by generative pre-training}.

\bibitem[{Radford et~al.(2019)Radford, Wu, Child, Luan et~al.}]{Radford2019LanguageMA}
Alec Radford, Jeff Wu, Rewon Child, David Luan, and 1 others. 2019.
\newblock \href {https://api.semanticscholar.org/CorpusID:160025533} {Language models are unsupervised multitask learners}.

\bibitem[{Srivastava et~al.(2023)Srivastava, Rastogi, Rao, Shoeb et~al.}]{Srivastava2023BeyondTI}
Aarohi Srivastava, Abhinav Rastogi, Abhishek Rao, Abu Awal~Md Shoeb, and 1 others. 2023.
\newblock \href {https://api.semanticscholar.org/CorpusID:271601672} {Beyond the imitation game: Quantifying and extrapolating the capabilities of language models}.
\newblock \emph{Trans. Mach. Learn. Res.}, 2023.

\bibitem[{Taylor et~al.(2022)Taylor, Kardas, Cucurull, Scialom et~al.}]{Taylor2022GalacticaAL}
Ross Taylor, Marcin Kardas, Guillem Cucurull, Thomas Scialom, and 1 others. 2022.
\newblock \href {https://api.semanticscholar.org/CorpusID:253553203} {Galactica: A large language model for science}.
\newblock \emph{ArXiv}, abs/2211.09085.

\bibitem[{Tian et~al.(2020)Tian, Yang, Liu, and Lv}]{Tian2020AnchiBERTAP}
Huishuang Tian, Kexin Yang, Dayiheng Liu, and Jiancheng Lv. 2020.
\newblock \href {https://api.semanticscholar.org/CorpusID:221879140} {Anchibert: A pre-trained model for ancient chinese language understanding and generation}.
\newblock \emph{2021 International Joint Conference on Neural Networks (IJCNN)}, pages 1--8.

\bibitem[{Touvron et~al.(2023)Touvron, Lavril, Izacard et~al.}]{Touvron2023LLaMAOA}
Hugo Touvron, Thibaut Lavril, Gautier Izacard, and 1 others. 2023.
\newblock \href {https://api.semanticscholar.org/CorpusID:257219404} {Llama: Open and efficient foundation language models}.
\newblock \emph{ArXiv}, abs/2302.13971.

\bibitem[{Vaswani et~al.(2017)Vaswani, Shazeer, Parmar, Uszkoreit, Jones, Gomez, Kaiser, and Polosukhin}]{Vaswani2017AttentionIA}
Ashish Vaswani, Noam~M. Shazeer, Niki Parmar, Jakob Uszkoreit, Llion Jones, Aidan~N. Gomez, Lukasz Kaiser, and Illia Polosukhin. 2017.
\newblock \href {https://api.semanticscholar.org/CorpusID:13756489} {Attention is all you need}.
\newblock In \emph{Neural Information Processing Systems}.

\bibitem[{Wang et~al.(2023{\natexlab{a}})Wang, Liu, Zhao, Shen et~al.}]{wang2023gujibert}
Dongbo Wang, Chang Liu, Zhixiao Zhao, Si~Shen, and 1 others. 2023{\natexlab{a}}.
\newblock Gujibert and gujigpt: Construction of intelligent information processing foundation language models for ancient texts.
\newblock \emph{arXiv preprint arXiv:2307.05354}.

\bibitem[{Wang et~al.(2022)Wang, Liu, Zhu, Liu et~al.}]{TSGL202206005}
Dongbo Wang, Chang Liu, Zihe Zhu, Jiangfeng Liu, and 1 others. 2022.
\newblock Sikubert and sikuroberta: Construction and application of pre-trained models for the siku quanshu in the field of digital humanities.
\newblock \emph{Library Tribune}, 42(06):31--43.

\bibitem[{Wang et~al.(2023{\natexlab{b}})Wang, Liu, Xi, Qiang, Zhao, Qin, and Liu}]{Wang2023HuaTuoTL}
Hao Wang, Chi-Liang Liu, Nuwa Xi, Zewen Qiang, Sendong Zhao, Bing Qin, and Ting Liu. 2023{\natexlab{b}}.
\newblock \href {https://api.semanticscholar.org/CorpusID:258170497} {Huatuo: Tuning llama model with chinese medical knowledge}.
\newblock \emph{ArXiv}, abs/2304.06975.

\bibitem[{Wang et~al.(2019)Wang, Wei, Guo, and Cheng}]{wang2019ancient}
Hongbin Wang, Haibing Wei, Jianyi Guo, and Liang Cheng. 2019.
\newblock Ancient chinese sentence segmentation based on bidirectional lstm+ crf model.
\newblock \emph{Journal of advanced computational intelligence and intelligent informatics}, 23(4):719--725.

\bibitem[{Xiong et~al.(2023)Xiong, Wang, Zhu, Zhao, Liu, Huang, Wang, and Shen}]{Xiong2023DoctorGLMFY}
Honglin Xiong, Sheng Wang, Yitao Zhu, Zihao Zhao, Yuxiao Liu, Linlin Huang, Qian Wang, and Dinggang Shen. 2023.
\newblock \href {https://api.semanticscholar.org/CorpusID:257912795} {Doctorglm: Fine-tuning your chinese doctor is not a herculean task}.
\newblock \emph{ArXiv}, abs/2304.01097.

\bibitem[{Yan et~al.(2016)Yan, Li, Hu, and Zhang}]{yan2016chinese}
Rui Yan, Cheng-Te Li, Xiaohua Hu, and Ming Zhang. 2016.
\newblock Chinese couplet generation with neural network structures.
\newblock In \emph{Proceedings of the 54th Annual Meeting of the Association for Computational Linguistics (Volume 1: Long Papers)}, pages 2347--2357.

\bibitem[{Yang et~al.(2023{\natexlab{a}})Yang, Xiao, Wang, Zhang et~al.}]{yang2023baichuan}
Aiyuan Yang, Bin Xiao, Bingning Wang, Borong Zhang, and 1 others. 2023{\natexlab{a}}.
\newblock Baichuan 2: Open large-scale language models.
\newblock \emph{arXiv preprint arXiv:2309.10305}.

\bibitem[{Yang et~al.(2024{\natexlab{a}})Yang, Yang, Zhang et~al.}]{yang2024qwen2}
An~Yang, Baosong Yang, Beichen Zhang, and 1 others. 2024{\natexlab{a}}.
\newblock Qwen2. 5 technical report.
\newblock \emph{arXiv preprint arXiv:2412.15115}.

\bibitem[{Yang et~al.(2024{\natexlab{b}})Yang, Zhao, Zhu, Zhou et~al.}]{yang2024zhongjing}
Songhua Yang, Hanjie Zhao, Senbin Zhu, Guangyu Zhou, and 1 others. 2024{\natexlab{b}}.
\newblock Zhongjing: Enhancing the chinese medical capabilities of large language model through expert feedback and real-world multi-turn dialogue.
\newblock In \emph{Proceedings of the AAAI Conference on Artificial Intelligence}, volume~38, pages 19368--19376.

\bibitem[{Yang et~al.(2023{\natexlab{b}})Yang, Tang, and Tam}]{yang2023investlmlargelanguagemodel}
Yi~Yang, Yixuan Tang, and Kar~Yan Tam. 2023{\natexlab{b}}.
\newblock \href {https://arxiv.org/abs/2309.13064} {Investlm: A large language model for investment using financial domain instruction tuning}.
\newblock \emph{Preprint}, arXiv:2309.13064.

\bibitem[{Yi et~al.(2017)Yi, Li, and Sun}]{yi2017generating}
Xiaoyuan Yi, Ruoyu Li, and Maosong Sun. 2017.
\newblock Generating chinese classical poems with rnn encoder-decoder.
\newblock In \emph{Chinese Computational Linguistics and Natural Language Processing Based on Naturally Annotated Big Data: 16th China National Conference, CCL 2017, and 5th International Symposium, NLP-NABD 2017, Nanjing, China, October 13-15, 2017, Proceedings 16}, pages 211--223. Springer.

\bibitem[{Yuan et~al.(2019)Yuan, Wang, Huang, and Li}]{XDTQ201903006}
Y~Yuan, D~Wang, S~Huang, and B~Li. 2019.
\newblock The comparative study of different tagging sets on entity extraction of classical books.
\newblock \emph{Data Analysis and Knowledge Discovery}, 3(03):57--65.

\bibitem[{Yue et~al.(2023)Yue, Chen, Wang, Li et~al.}]{Yue2023DISCLawLLMFL}
Shengbin Yue, Wei Chen, Siyuan Wang, Bingxuan Li, and 1 others. 2023.
\newblock \href {https://api.semanticscholar.org/CorpusID:262064568} {Disc-lawllm: Fine-tuning large language models for intelligent legal services}.
\newblock \emph{ArXiv}, abs/2309.11325.

\bibitem[{Zeng et~al.(2024)Zeng, Xu, Wang, Zhang, Yin, Rojas et~al.}]{Zeng2024ChatGLMAF}
Team Glm~Aohan Zeng, Bin Xu, Bowen Wang, Chenhui Zhang, Da~Yin, Diego Rojas, and 1 others. 2024.
\newblock \href {https://api.semanticscholar.org/CorpusID:270562306} {Chatglm: A family of large language models from glm-130b to glm-4 all tools}.
\newblock \emph{ArXiv}, abs/2406.12793.

\bibitem[{Zhang et~al.(2024)Zhang, Yang, Liu, and Huang}]{TSGL202410011}
Jundong Zhang, Songhua Yang, Jiangfeng Liu, and Qi~Huang. 2024.
\newblock Aigc empowering the revitalization of ancient books on traditional chinese medicine:building the huang-di large language model.
\newblock \emph{Library Tribune}, 44(10):103--112.

\bibitem[{Zhang et~al.(2023{\natexlab{a}})Zhang, Yang, and Xu}]{Zhang2023XuanYuan2A}
Xuanyu Zhang, Qing Yang, and Dongliang Xu. 2023{\natexlab{a}}.
\newblock \href {https://api.semanticscholar.org/CorpusID:258833440} {Xuanyuan 2.0: A large chinese financial chat model with hundreds of billions parameters}.
\newblock \emph{Proceedings of the 32nd ACM International Conference on Information and Knowledge Management}.

\bibitem[{Zhang et~al.(2023{\natexlab{b}})Zhang, Deng, Zhang, Wang, and Gong}]{zhang2023comparative}
Yiqin Zhang, Sanhong Deng, Qi~Zhang, Dongbo Wang, and Hongcun Gong. 2023{\natexlab{b}}.
\newblock Comparative analysis of language models for linguistic examination of ancient chinese classics: A case study of zuozhuan corpus.
\newblock In \emph{2023 International Conference on Asian Language Processing (IALP)}, pages 154--161. IEEE.

\end{thebibliography}


\appendix

\section{Categories of part-of-speech tagging}
\label{sec:appendix1}
The part-of-speech tagging includes the following 17 categories:
nr – Proper Noun (Person), v – Verb, n – Noun, r – Pronoun, w – Punctuation, c – Conjunction, p – Preposition, d – Adverb, t – Time word, y – Modal Particle, u – Auxiliary word, m – Numeral, a – Adjective, f – Locality Word, ns – Proper Noun (Place), j – Abbreviation, q – Quantifier.

\section{Example of instruction construction}
\label{sec:appendixa2}
(1) Initialization instruction template: 

instruction = "Extract entities from the following Classical Chinese paragraph, and list characters, place, time, and official titles. Return the result in the following format: 'characters': [...] , 'place': [...] , 'time': [...] , 'official positions': [...]."

(2) Instruction for task expansion:

prompt = "Based on the following example instruction, generate additional instructions for the Named Entity Recognition (NER) task. These instructions should ask the model to extract four types of entities include characters, place, time, and official positions from a given Classical Chinese paragraph, and return the results in structured format. The generated instructions should be diverse, natural, consistent with Chinese linguistic habits, and should clearly specify the task goal and output format. Example instruction: \{instruction\}"

(3) Reverse reasoning:

prompt = "Based on the given input content and the corresponding output, generate some clear and reasonable task instructions. The instructions should effectively guide the model to extract four categories of named entities include characters, place, time, and official positions from Classical Chinese text and return them in a structured format consistent with the provided output. Please ensure linguistic diversity while maintaining semantic accuracy.

Reference example:

Input: Extract characters, place, time, and official positions from the following ancient texts: 
\begin{CJK}{UTF8}{gbsn}侍御史赵承庆见之，叹曰：御史言及此，天下福也。\end{CJK}

Output: 'characters': ['\begin{CJK}{UTF8}{gbsn}赵承庆\end{CJK}'], 'place': [], 'time': [], 'official positions': ['\begin{CJK}{UTF8}{gbsn}御史\end{CJK}']"

Answer: 

1. Please conduct entity analysis of the following classical Chinese to identify the full names, geographical names, historical time nodes and official information of all characters, and classify them in JSON format. If there is no corresponding entity for a category, leave the list empty.

2. You need to find out four types of information from the following ancient materials: the name of the person, the specific place, the dynasty/year, and the official position. Please according to the sample format arrangement results: {' characters' : [], 'place': [], 'time': [], 'official positions': []}"

...

(4)Filtering generated instructions:

Accept Example :

Input: Perform named entity recognition on the given Classical Chinese text. Extract characters, place, time, and official positions, and return them in the following format: 'characters': [...], 'place': [...], 'time': [...], 'official positions': [...]. Classical Chinese text: \begin{CJK}{UTF8}{gbsn}天成初，移镇许州，加同平章事。\end{CJK}

Output:'characters': [], 'place': ['\begin{CJK}{UTF8}{gbsn}许州\end{CJK}'], 'time': ['\begin{CJK}{UTF8}{gbsn}天成初\end{CJK}'], 'official positions': ['\begin{CJK}{UTF8}{gbsn}同平章事\end{CJK}']

Reject Example:

Input: Perform named entity recognition on the given Classical Chinese text: \begin{CJK}{UTF8}{gbsn}天成初，移镇许州，加同平章事。\end{CJK}

Output: \begin{CJK}{UTF8}{gbsn}天成初：时间，许州：地点，同平章事：官职。\end{CJK}

\section{Case of word explanation task}
\label{sec:appendix3}
\begin{table}[h!]
	\centering
	\begin{tabular}{p{2.5cm}p{4.6cm}}
		\hline
		\textbf{Case 2} & \begin{CJK}{UTF8}{gbsn}Explain the meaning of "晕红" in the line "一片晕红才著雨，几丝柔绿乍和烟" from "浣溪沙·谁道飘零不可怜".\end{CJK} \\
		\hline
		\textbf{Ground Truth} & \begin{CJK}{UTF8}{gbsn}形容海棠花的色泽。\end{CJK} Describes the color of begonia flowers. \\
		\hline
		\textbf{WenyanGPT} & \begin{CJK}{UTF8}{gbsn}形容海棠花色泽鲜艳而浓丽。\end{CJK} Describes the bright and rich color of the begonia flower. \\
		Deepseek-V3 & \begin{CJK}{UTF8}{gbsn}指雨后花朵泛起的朦胧红晕。\end{CJK} It refers to the hazy red glow that appears on flowers after the rain. \\
		GPT-4o & \begin{CJK}{UTF8}{gbsn}指花瓣在雨水滋润下呈现出的一种柔和而略带湿润光泽的红色。\end{CJK} Refers to the soft and slightly moist red color that the petals appear to have when nourished by rain. \\
		Qwen2.5-7B-Instruct & \begin{CJK}{UTF8}{gbsn}指花瓣上晕染开的淡红色，适合颜色柔和、渐变，类似于涂抹或被雨水滋润后散开的红晕。\end{CJK} Refers to the light red color that spreads across the petals. It is suitable for soft and gradual colors, similar to the blush that spreads after being smeared or moistened by rain. \\
		Xunzi-Qwen1.5-7B-Chat & \begin{CJK}{UTF8}{gbsn}指花瓣上的红色晕染，因为花瓣在雨中被淋湿，所以红色晕染更加明显。\end{CJK} Refers to the red smudge on the flower petals. Because the petals got wet in the rain, the red smudge is more obvious. \\
		\hline
	\end{tabular}
	\caption{Response examples for the word explanation task from different LLMs.}
	\label{tab:10}
\end{table}

For the \textbf{word explanation} task, WenyanGPT's response is the closest to the Ground Truth. Although WenyanGPT adds some modifiers, it maintains simplicity and refinement in describing the color, which matches the tone of the poem. Deepseek-V3, GPT-4o, Qwen2.5-7B-Instruct, and Xunzi-Qwen1.5-7B-Chat do not specify the subject as "\begin{CJK}{UTF8}{gbsn}海棠花\end{CJK}" (begonia flowers) and instead over-explained the moistness of the petals and the process of spreading color, deviating from the standard, concise description of "\begin{CJK}{UTF8}{gbsn}晕红\end{CJK}" (blushing red). Specifically, the responses from Qwen2.5-7B-Instruct and Xunzi-Qwen1.5-7B-Chat extend the impact of the external environment, which do not fully align with the original text's expression. This over-expansion shows that these LLMs fail to capture the intended simplicity and depth of the original description.
\end{document}